\documentclass[journal]{IEEEtran}
\usepackage{comment}

\usepackage{cite}
\usepackage{subfig}
\usepackage{tikz}
\usepackage{booktabs}
\usepackage{tikz}
  \usetikzlibrary{arrows}
  \usetikzlibrary{automata}
  \usetikzlibrary{spy}
\usepackage{pgfplots}
  \pgfplotsset{width=7cm,compat=1.8}

\usepackage{amsmath,amssymb}

\usepackage{soul}

\usepackage{placeins}

\usepackage[numbers]{natbib}
\usepackage{mathrsfs}
\usepackage{xcolor}
\begin{document}
%
\title{Tree species classification from hyperspectral data using graph-regularized neural networks}
%
%
%

\author{Debmita Bandyopadhyay,
        Subhadip Mukherjee, James Ball, Grégoire Vincent, David A. Coomes, 
        and Carola-Bibiane Schönlieb 
\thanks{The work was supported by GCRF (EPSRC) grant EP/T003553/1, the Isaac Newton Trust fund (21.23(f)) and the  Franklina Foundation. }
\thanks{DB, JB, and DC are members of the Department
 of Plant Sciences and Conservation Research Institute, University of Cambridge, UK. SM is with the Department of Computer Science, University of Bath, UK. GV is with AMAP, Univ Montpellier, CIRAD, CNRS, INRAE, IRD, Montpellier, France.  CBS is with the Department of Applied Mathematics and Theoretical Physics, University of Cambridge, UK, email: cbs31@cam.ac.uk. }}
\maketitle

\begin{abstract}
We propose a novel graph-regularized neural network (GRNN) algorithm for tree species classification. The proposed algorithm encompasses superpixel-based segmentation for graph construction, a pixel-wise neural network classifier, and the label propagation technique to generate an accurate and realistic (emulating tree crowns) classification map on a sparsely annotated data set. GRNN outperforms several state-of-the-art techniques not only for the standard Indian Pines HSI but also achieves a high classification accuracy (approx. 92\%) on a new HSI data set collected over the heterogeneous forests of French Guiana (FG) when less than 1\% of the pixels are labeled. 
We further show that GRNN is competitive with the state-of-the-art semi-supervised methods and exhibits a small deviation in accuracy for different numbers of training samples and over repeated trials with randomly sampled labeled pixels for training.  

\end{abstract}


%
\IEEEpeerreviewmaketitle

\section{Introduction}
\IEEEPARstart{T}{ree} species classification of forests is important as several commercially important and vulnerable species are fast depleting. Using non-invasive technologies, like aerial hyperspectral sensors, tree species can be distinguished at scale even with very similar spectral curves. Most of the studies on tree-species classification using HSI data have either employed pixel-level supervised classification methods (e.g., support vector machines (SVMs), linear discriminant analysis, or random forests)  or utilized crown cover boundaries to provide realistic classification maps \cite{SVM_2004TGRS,RF_breiman_2001,MDPI_LDA}. However, these techniques rely heavily on apriori information with limited capability to extract spectral-spatial features from the raw data. Deep learning-based models like CNNs, RNNs and GANs can learn from the spatial texture information \cite{2021_Safari_DL} but have a high probability of misalignment between the network's objectives of achieving high classification accuracy and providing a realistic tree species classified image. The resultant image is highly accurate but unrealistically classified, especially for the unlabeled pixels in the image \cite{zhang2023_deeplearning_tree-classification}.\\
For tropical forests, apart from having spectral similarity between species but heterogeneity in species distribution, there is an additional challenge of the limited availability of manual labels owing to the inaccessibility of these regions and the cost of establishing forest plots. Semi-supervised graph learning can tackle the problem of sparsely annotated data labels and extract useful information from labeled and unlabeled pixels in an image \cite{zhu2005ssl_smalllabels}. Hence, they are a preferred choice over deep learning methods as the latter depends on high proportions of annotated data. Besides, hyperspectral data involves dealing with high-resolution and high-dimensional spectral-spatial data sets. Various feature extraction methods like the image fusion and recursive filter by \citet{kang2013_IFRF}, utilization of local binary patterns by \citet{li2015_LBP}, local matrix representation by \citet{Fang2018_LCMR} and the hyper-manifold simple linear iterative clustering combined with the graph learning technique by \citet{Sellars_2020TGRS} have been shown to successfully address this high-dimensionality problem.  Additionally, the superpixel-based classification emulates the tree crown, providing a realistic classification map. 
In summary, the superpixel-based semi-supervised method deals with the (i) limited ground information challenge by clustering pixels with similar features, (ii) maintains a competitive accuracy alongside the state-of-the-art techniques, and (iii) provides a realistic (tree-crown scale) classification map. However, to preserve the pixel-based information and augment label propagation to unlabeled pixels, a neural net (NN)-based classifier is also required owing to their expressive capacity and the ability to generate an accurate pixel-level classification.\\
\noindent\textbf{Our contributions}: We propose a novel graph-regularized neural network (GRNN) classifier, which is built on the following three mechanisms: (1) First, to combine pixel-based classification and superpixel feature-based learning to extract maximum information from both labeled and unlabeled pixels in the HSI, (2) Second, to generate an accurate and realistic classification map by jointly optimizing the accuracy of a pixel-level NN and a superpixel-level graph energy potential that seeks to promote smoothness in the predicted classification map, and (3) third, the NN output predictions (exceeding certain confidence) are used to augment the ground-truth labels, on which we apply the classical semi-supervised label propagation method \cite{zhou2003_labelpropagation} using the superpixel graph to generate the final classification map. To the best of our knowledge, GRNN is the first framework that combines neural networks with superpixel graphs for the semi-supervised classification of tree species. 
For a data set where the tree species have similar spectral responses, the proposed GRNN algorithm (i) can accurately classify the top ten most commonly occurring species, (ii) provide a smooth superpixel-based classification map with a realistic view of the tree crowns, and (iii) works well even with $<1\%$ of the pixels being annotated. Details of GRNN and our experimental findings are reported in Sections \ref{sec:methods} and \ref{sec:exp}, respectively.

\section{Proposed method}
\label{sec:methods}
\begin{figure}[!h]
    \centering
    \includegraphics[width=3.5in]{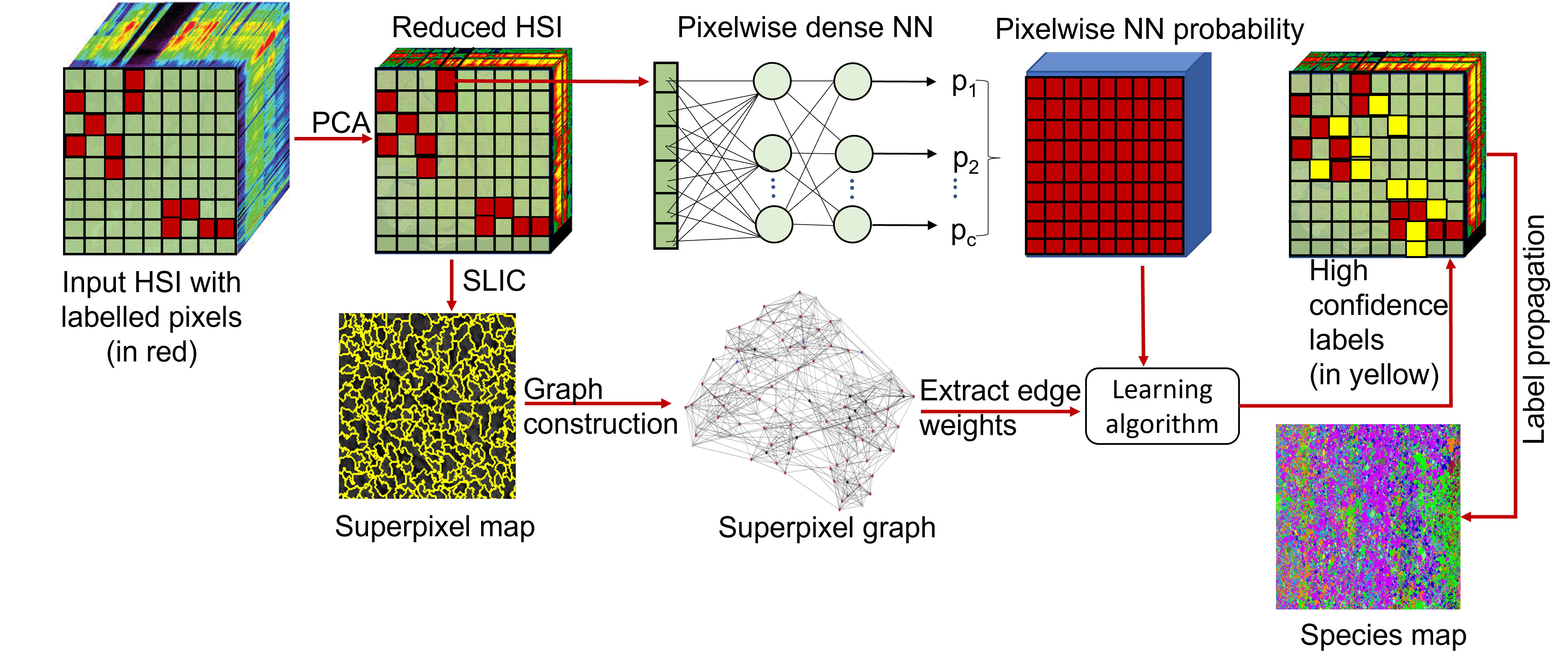}
    \caption{Framework for the proposed method, GRNN.}
    \label{fig:workflow}
\end{figure}

The key idea behind the GRNN classification approach is to use a pixelwise fully-connected NN classifier together with a superpixel graph to learn effectively from a few labeled pixels and a large number of unlabeled pixels. The output probability distribution of the NN is regularized using a graph energy function that smooths variation of the NN output. The details of the GRNN algorithm are explained in the following, and a schematic of the entire pipeline is shown in Figure \ref{fig:workflow}. 

The given HSI $X\in \mathbb{R}^{H\times W\times B}$, having spatial dimensions $H\times W$ and $B$ spectral bands, is first reduced via principal component analysis (PCA) to $x\in \mathbb{R}^{H\times W\times b}$, where $b\ll B$. The number of bands $b$ is chosen such that 99.90\% of variance in the HSI data is preserved. The labeled and unlabeled pixels in the given HSI are denoted by $\mathcal{L}$ and $\mathcal{U}$, respectively. 
\noindent Following PCA, we apply the \textit{simple linear iterative clustering} (SLIC) algorithm \cite{Achanta2012_SLIC} to the first principal component of $X$ to extract $N$ superpixels, denoted as $\left\{\mathcal{S}_k\right\}_{k=1}^{N}$, which are subsequently utilized to construct a graph. An essential step for superpixel graph construction is \textit{feature extraction}, for which we follow the procedure outlined in \cite[Sec. III.B]{Sellars_2020TGRS}. The nodes of the graph represent the superpixels, and the adjacency weights $W_{k\ell}$ capture pairwise similarity between superpixels $\mathcal{S}_k$ and $\mathcal{S}_{\ell}$ in terms of the extracted features. We follow the same procedure as reported in \cite{Sellars_2020TGRS} for constructing the graph adjacency matrix $W\in \mathbb{R}^{N\times N}$. Small adjacency weights are set to $0$ by retaining only the top $N_s$ elements in each row of $W$. This thresholding step sparsifies $W$ and leads to efficient training. We choose $N_s=20$ in our experiments, i.e., each node can have at most 20 neighboring nodes. A superpixel $\mathcal{S}_k$ is said to be labeled if it contains at least one labeled pixel in it, i.e., $\mathcal{S}_k\cap \mathcal{L}\neq \emptyset$, and the set of all labeled superpixels is denoted as $\mathcal{L}^{\text{sp}}$. For a $c$-class classification problem, let $y_\mathbf{i}\in\left\{1,2,\cdots,c\right\}$ be the true label of any pixel $\mathbf{i}=(i_1,i_2) \in \mathcal{L}$. The one-hot encoding of $y_\mathbf{i}$ is denoted by $Y_\mathbf{i} \in\mathbb{R}^c$, and is defined as follows:
\begin{equation}
Y_{\mathbf{i}}(q)=\begin{cases}
1 & \text{ if $y_{\mathbf{i}}=q$, where $q\in \left\{1,2,\cdots,c\right\}$, and} \\
0 & \text{ if $y_{\mathbf{i}}\neq q$.}
\end{cases}
\label{eq:onehot_enc}
\end{equation}
For any labeled superpixel $\mathcal{S}_k\in \mathcal{L}^{\text{sp}}$, we compute two labels, referred to as the \textit{soft} and the \textit{hard} labels, respectively. The soft label $t_{\mathcal{S}_k}$ represents the relative frequency of different classes within it, and is computed as $\displaystyle t_{\mathcal{S}_k}=\frac{1}{\left|\mathcal{S}_k\cap \mathcal{L}\right|}\sum_{\mathbf{j}\in \mathcal{S}_k \cap \mathcal{L}}Y_{\mathbf{j}}$.
Note that for any $\mathcal{S}_k\in \mathcal{L}^{\text{sp}}$, the soft label $t_{\mathcal{S}_k}\in \Delta_c$, where $\Delta_c$ is the probability simplex in $\mathbb{R}^c$. The hard label $T_{\mathcal{S}_k}$ corresponding to $\mathcal{S}_k$ represents the most frequent class in $\mathcal{S}_k$ using a one-hot encoding. Specifically, $T_{\mathcal{S}_k}\in\mathbb{R}^c$ is a $c$-length vector whose $q_{*}^{\text{th}}$ element is 1, and every other element is 0, where $q_{*}=\underset{q\in\{1,2,\cdots,c\}}{\arg \max}t_{\mathcal{S}_k}(q)$. For an unlabeled superpixel $\mathcal{S}_k\notin \mathcal{L}^{\text{sp}}$, both $t_{\mathcal{S}_k}$ and $T_{\mathcal{S}_k}$ are defined as $c$-length vectors with all zeros. Let, $T^{(\mathcal{L})}$ be the $N\times c$ matrix whose $k^{\text{th}}$ row is $T_{\mathcal{S}_k}$. The superscript $\mathcal{L}$ is used to indicate explicitly that $T$ depends on the underlying set of labeled pixels $\mathcal{L}$.  

Besides the superpixel graph, the other key component of GRNN is a pixel-wise, fully-connected NN-based classifier $\phi_{\theta}:\mathbb{R}^b\mapsto \Delta_c$ with parameters $\theta$. The NN has two intermediate layers with leaky-ReLU activations (with negative slope 0.1)\footnote{\footnotesize{$\text{leaky-relu}(u)=\max(0,u)+\gamma\min(0,u)$, where $\gamma=$ negative slope.}} and a softmax layer in the end. The output of the NN gives a probability map over $c$ classes, i.e., for any pixel location $\mathbf{i}\in\mathbb{R}^2$, $\phi_{\theta}(x_{\mathbf{i}})\in \Delta_c$ represents the predicted probabilities of the pixel belonging to any of the $c$ classes. The predicted probability map for a superpixel is computed by averaging the probability maps of all the pixels within the given superpixel: $\phi_{\mathcal{S}_k}=\frac{1}{\left|\mathcal{S}_k\right|}\sum_{\mathbf{j}\in \mathcal{S}_k}\phi_{\theta}(x_{\mathbf{j}})$.
The overall training loss seeks to achieve the following objectives simultaneously: (i) minimize classification errors, both for the pixels and the superpixels for which the ground-truth labels are available, (ii) ensure that the predicted superpixel probability map varies smoothly over the graph, (iii) minimize the standard deviation in the predicted pixelwise probability maps within each superpixel, and (iv) discourage uniform labeling of the unlabeled superpixels, i.e., make sure that all unlabeled superpixels do not end up with the same class and that all classes are represented overall. To accomplish these objectives, the training problem is formulated as  
\begin{align}
    \underset{\theta}{\min}&\left[ \sum_{\mathbf{j}\in\mathcal{L}}\mathcal{D}_1\left(Y_{\mathbf{j}},\phi_{\theta}(x_{\mathbf{j}})\right)+\lambda_{\text{spc}}\sum_{k:\mathcal{S}_k\in\mathcal{L}^{\text{sp}}}\mathcal{D}_2\left(t_{\mathcal{S}_k},\phi_{\mathcal{S}_k}\right)\nonumber\right.\\
    &+\lambda_{\text{g}} \sum_{\mathcal{S}_k,\mathcal{S}_{\ell}}\left\|\frac{\phi_{\mathcal{S}_k}}{\sqrt{d_k}}-\frac{\phi_{\mathcal{S}_{\ell}}}{\sqrt{d_{\ell}}}\right\|_2^2+\lambda_{\text{v}}\sum_{\mathcal{S}_k}\text{var}\left\{\phi_{\theta}(x_{\mathbf{j}})\right\}_{\mathbf{j}\in\mathcal{S}_k}\nonumber\\
    &\left.-\lambda_{\text{en}}\mathcal{H}\left(\frac{1}{N}\sum_{k=1}^{N}\phi_{\mathcal{S}_k}\right)\right].
    \label{eq:training_loss}
\end{align}
Here, $d_k=\sum_{\ell:\ell\neq k}W_{k \ell}$ denotes the degree of node-$k$ of the superpixel graph, $\mathcal{D}_1$ and $\mathcal{D}_2$ measure pixel- and superpixel-wise classification losses, respectively, and $\mathcal{H}:\Delta_c\mapsto \mathbb{R}$ denotes the entropy functional. The first two terms in \eqref{eq:training_loss} promote high classification accuracy on the labeled pixels and superpixels, respectively. We choose $\mathcal{D}_1$ to be the cross-entropy loss, while $\mathcal{D}_2$ is taken as the standard squared Euclidean distance. The third term is a graph energy function that enforces a smoothness prior on the predicted superpixel probability map, ensuring that it does not vary abruptly as a function defined on the superpixel graph. The fourth term in the training loss seeks to minimize the variance of prediction probabilities for all pixels inside each superpixel, which promotes high intra-superpixel similarity of the prediction. The final negative entropy term penalizes uniform labeling of the unlabeled superpixels, thereby encouraging high inter-superpixel variability of the predicted classification map (which is well-suited for the highly heterogeneous French Guiana data considered in Sec. \ref{sec:exp}). The $\lambda$ parameters in the overall training loss trade off different components of the loss function and need to be selected optimally. The training problem \eqref{eq:training_loss} is solved iteratively using the Adam optimizer \cite{adam_kingma}, the de facto standard for training NNs, with a learning rate of 10\textsuperscript{--3} and $(\beta_1,\beta_2)=(0.9,0.999)$. 

The trained NN, although optimized simultaneously for accuracy and low intra-superpixel variability of classification, can still produce a classification map with differently labeled pixels inside a given superpixel. Such a prediction might be physically unrealistic since the superpixels are generated based on spectral similarity and any given superpixel should ideally belong to a single tree species. To avoid this, the final step of GRNN generates a smooth label map by applying the classical label propagation algorithm for semi-supervised learning on graphs \cite{zhou2003_labelpropagation}, after combining the NN prediction with the ground-truth labels. The label propagation step is carried out by augmenting the given ground-truth labels with the labels predicted by the NN with high confidence. The set of pixels for which the trained NN classifies with confidence exceeding a given threshold $\tau\in(0,1)$ is defined as
\begin{equation}
    \mathcal{I} = \left\{\mathbf{i}:\max\left(\phi_{\theta^*}(x_{\mathbf{i}})\right)\geq \tau\right\},
    \label{eq:threshold}
\end{equation}
where $\theta^*$ is the optimal NN parameters obtained by solving \eqref{eq:training_loss}. In other words, if the probability of the most likely class is greater than $\tau$ (we set $\tau=0.4$) then the corresponding predicted label is added to the existing ground-truth labels. 
The smoothed superpixel probability map is generated by 
\begin{equation}
T^*=\left(\text{Id}-\alpha\,D^{-\frac{1}{2}}WD^{\frac{1}{2}}\right)^{-1}T^{(\mathcal{M})},  
\label{eq:label_smoothing}
\end{equation}
where $\text{Id}$ denotes the $N\times N$ identity matrix, $T^{(\mathcal{M})}$ is the $N\times c$ one-hot superpixel probability matrix computed using the augmented set of labeled pixels $\mathcal{M}=\mathcal{L}\cup\mathcal{I}$, and $W\in \mathbb{R}^{N\times N}$ and $D\in \mathbb{R}^{N\times N}$ are the graph adjacency and the (diagonal) degree matrices, respectively. For any superpixel $\mathcal{S}_k$, the final classification label is produced as $q_k = \underset{q\in\{1,2,\cdots,c\}}{\arg \max}T^*_{\mathcal{S}_k}(q)$, where $T^*_{\mathcal{S}_k}$ is the $k^{\text{th}}$ row of $T^*$. All pixels within a superpixel are assigned the same class label as their parent superpixel. 

\section{Experiments and results}
\label{sec:exp}
\subsubsection{Data description/set up}

The widely used Indian Pines (IP) HSI benchmark data set is used to compare the proposed method with the existing state-of-the-art (SoTA) classification techniques comprising 16 classes. The data was collected over an agricultural site by the airborne visible/infrared imaging spectrometer (AVIRIS) sensor. We chose the IP data for performance comparison and benchmarking since all the competing SoTA methods were previously applied to this data and optimized for their respective best performances.

To check the robustness of GRNN, we apply it to a new data set collected with Hyspex VNIR 1600 and Hyspex SWIR 384 sensor-mounted alongside Riegl scanner. The study site is Paracou Research Station (5$^{\circ}$18'N, 52$^{\circ}$55'W) in French Guiana (FG), which is in the tropical rainforest. The data set comprises  76 tree species \cite{FrenchGuana_JSTARS2021}. The ground truth labels were manually confirmed from the field survey, and the crowns were manually segmented. Only the top 10 most dominant species (comprising 6\% of the total number of pixels) are considered for classification, and the other species were excluded (not assigned any label). 
Our experiment is performed in two stages. First, we develop and test the algorithm on the benchmark IP HSI data and compare the overall accuracy (OA) and the Cohen kappa coefficient ($\kappa$) with competing classification methods. The mean and standard deviation of classification accuracy were evaluated over ten random trials for this benchmark HSI data. Subsequently, we test GRNN on the large, highly heterogeneous FG HSI data with very limited ground-truth labels. Among the labeled pixels, the test and training samples were randomly chosen to estimate the OA and the $\kappa$ score. 
\subsubsection{Hyperparameter selection}
GRNN involves two sets of hyperparameters, one for graph construction and the other for training. The optimal selection of these hyperparameters is crucial for the success of GRNN. The specific values chosen for the hyperparameters corresponding to two different data sets are reported in Table \ref{tab:hyperparams}. We chose these values based on cross-validation on a small validation set kept aside from the labeled data used for training. A more principled approach to learning these hyperparameters from the data would be highly desirable, and we leave it as future work.

\begin{table}[h]
    \centering
    \caption{\small{The hyperparameters for graph construction and GRNN (chosen through cross-validation). $h$, $\beta$, $\sigma_s$ and $\sigma_l$ are as defined in (5), (8), and (9) of \cite{Sellars_2020TGRS}. $\xi = $ connectivity defining superpixel adjacency. $n_{\theta} =$ total number of NN parameters. $N = $ number of superpixels extracted using SLIC, the $\lambda$ parameters control different components of the training objective \eqref{eq:training_loss}, $\alpha =$ label-smoothing parameter in \eqref{eq:label_smoothing}, $\tau =$ threshold in \eqref{eq:threshold}, and $n_{\text{iter}}=$ \# training iterations.}}
    \begin{tabular}{cccccccc}
        \hline
        \multicolumn{8}{c}{\textbf{superpixel graph construction and NN parameters}}\\
        & $h$ & $\beta$ & $\sigma_s$ & $\sigma_{l}$ & $\xi$ & $n_{\theta}$ &\\
        \hline
        Indian Pines & 15.0 & 0.9 & 2.0 & 1.0 & 8 & 73888 & \\
        French Guiana & 15.0 & 0.5 & 5.0 & 40.0 & 8 & 369778 & \\
        \midrule
        \multicolumn{8}{c}{\textbf{GRNN parameters}}\\
        & $N$ & $\lambda_{\text{spc}}$ & $\lambda_{\text{g}}$ &  $\lambda_{\text{v}}$ & $\lambda_{\text{en}}$ & $\alpha$ & $n_{\text{iter}}$    \\
        \hline 
   Indian Pines  &  1200  & 0.15 & 10\textsuperscript{5} & 2.0 & 20.0 & 0.5  & 500\\
   French Guiana  &  5000  & 0.1 & 0.2 & 0.1 & 20.0 & 0.5 & 400\\
         \hline
    \end{tabular}
    \label{tab:hyperparams}
\end{table}

\subsubsection{Comparison with different methods}
We compare and benchmark the performance of the proposed GRNN framework with respect to several SoTA methods, namely superpixel graph learning (SGL) \cite{Sellars_2020TGRS}, the local covariance matrix representation (LCMR) \cite{Fang2018_LCMR}, superpixel-based classification via multiple kernels (SC-MK) \cite{fang2015_SCMK}, edge-preserving filter (EPF) \cite{kang2013_EPF}, local binary pattern (LBP) method \cite{li2015_LBP}, the image fusion and recursive filtering (IFRF) \cite{kang2013_IFRF}, and the traditional SVM \cite{SVM_2004TGRS}. 

\begin{table*}[h]
\caption{\small{Comparison of different methods on Indian Pines in terms of overall accuracy and the $\kappa$ score, corresponding to ten samples per class. GRNN performed the best (highlighted in boldface).}}
\label{table_example}
\centering
\begin{tabular}{ccccccccc}
\hline
& GRNN (Ours) & SGL & LCMR & SC-MK & EPF & LBP & IFRF & SVM\\[1.2ex]
\hline
OA (\%) & $\mathbf{96.3 \pm 0.8}$ & $90.7 \pm 2.2$ & $82.7 \pm 3.1$& $80.7 \pm 2.5$ & $67.3 \pm 3.2$ & $78.9 \pm 2.7$ &$80.3 \pm 1.8$ &$53.1 \pm 1.9$\\ [1.2ex]
$\kappa$& $\mathbf{0.96 \pm 0.01}$ & $0.88 \pm 0.03 $ & $0.81 \pm 0.03$ & $0.77 \pm 0.03$ & $0.65 \pm 0.02$ & $0.78 \pm 0.02$ & $0.78 \pm 0.04$ & $0.48 \pm 0.02$\\ 
\hline
\end{tabular}
\label{tab:method comparision}
\end{table*}

\begin{figure*}[h]
\centering
\subfloat[\small{RGB}]{
\includegraphics[width=0.15\linewidth]{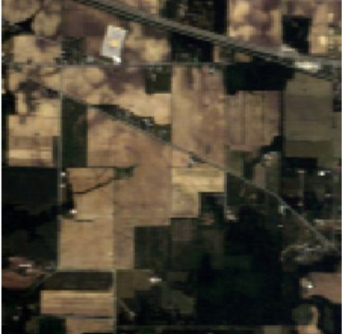}}
\subfloat[\small{ground-truth}]{
\includegraphics[width=0.15\linewidth]{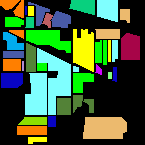}}
\subfloat[\small{SVM}]{
\includegraphics[width=0.15\linewidth]{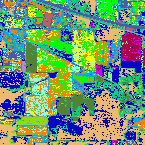}}
\subfloat[\small{SGL}]{
\includegraphics[width=0.15\linewidth]{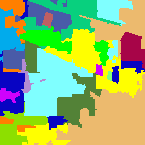}}
\subfloat[\small{GRNN}]{
\includegraphics[width=0.15\linewidth]{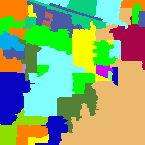}}
\subfloat{
\includegraphics[width=0.11\linewidth]{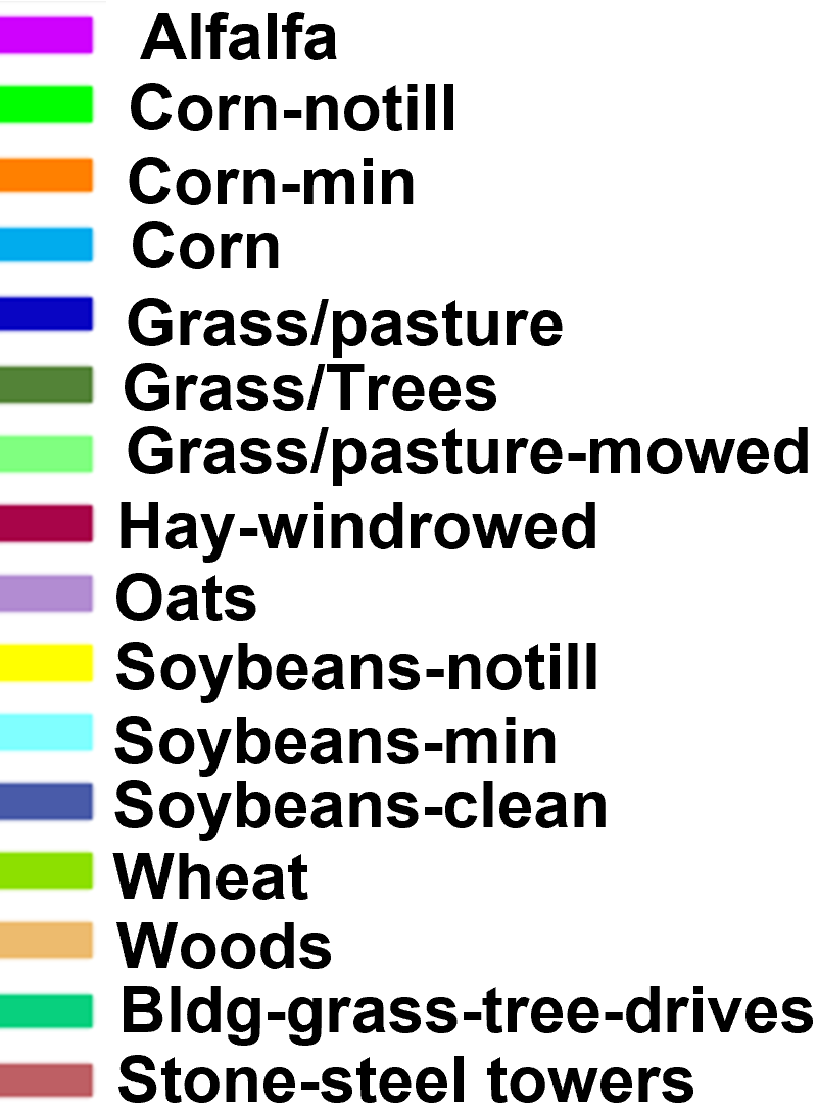}}
\caption{\small{Comparison of SVM (supervised), SGL (semi-supervised), and the proposed GRNN  classification maps on Indian Pines with ten samples per class. Qualitatively, GRNN produces a smoothed classification map with a higher accuracy as reported in Table \ref{tab:method comparision}.}}
\label{fig:pines_images}
\end{figure*}
GRNN performs both qualitatively (Figure \ref{fig:pines_images}) and quantitatively (Table \ref{tab:method comparision}) better than the existing classifiers for the Indian Pines data set. Specifically, the difference in $\kappa$ from the nearest competitive algorithm SGL (which is also based on superpixel segmentation and graph-based semi-supervised label propagation, but does not use any NN) is about 0.08. The key difference between SGL and GRNN is the NN-assisted graph learning which utilizes both pixel- and superpixel-level information for efficient semi-supervised classification while not compromising on pixelwise classification accuracy. The high-confidence NN predictions are used to augment the scarce ground-truth labels, and a visibly smooth classification is generated using the label propagation approach \cite{zhou2003_labelpropagation,Sellars_2020TGRS}. 

To study the effect of the number of labeled samples on the overall accuracy, we compared GRNN with one of the best-performing semi-supervised algorithms (namely, SGL) and the classical supervised method of SVM (see Figure 1 in the supplementary document). We found that the OA in all three methods consistently increases as the number of samples increases from three to fifteen. However, GRNN does not exhibit a noticeable change in accuracy (varies between 95\% and 96\%) with a reduction in the number of training samples. The deviation in OA for GRNN is also significantly smaller in comparison with the other methods. The standard deviation (SD) for GRNN varies from 0.58 to 0.80, whereas the SD is as high as 3.85 for SGL and 5.30 for SVM. 

\noindent Subsequently, we extend the study to a new data set from French Guiana for which the data has very limited labeled information (with a smaller fraction of labeled pixels than the standard IP data). The FG data set has highly heterogeneous classes which change as one moves from one crown to another. Further, there is an imbalance in class representation, and we, therefore, select the ten most commonly occurring classes to train the model and generate the classification map. The GRNN algorithm, after hyperparameter tuning (Table \ref{tab:hyperparams}), is trained using 10\%, 30\%, 50\% and 70\% of the labeled data. The 10\% of the labeled data constitutes $\leq$ 1\% of the total number of pixels in the HSI. Even with such few labeled pixels, the algorithm achieves a $\kappa$ of $0.79$. As the training sample percentage increases, $\kappa$ improves up to 0.89. However, the performance plateaus beyond 50\% training data usage. The experiments indicate that GRNN is particularly suitable when very few pixels are labeled. We compare the performance of GRNN with SVM (as a baseline pixelwise classifier) and the SGL method on the FG data (Table III). Notably, SGL (the closest competitor of GRNN on IP data) and SVM work solely at the superpixel- and pixel levels, respectively. The comparison demonstrates the shortcomings of methods that work purely on either the pixels or the superpixels and underscores the efficacy of a method such as GRNN that combines pixel-level classification with a superpixel-level regularization. Such a hybrid approach has the potential to achieve a good balance between high overall pixel-wise classification accuracy and the ability to accurately capture the tree crowns, as indicated by the final classification maps shown in Fig. \ref{fig:fg_images}. The superpixel graph regularization also prevents the NN from overfitting when trained on a limited number of labeled pixels and leads to better generalization.  


The success of GRNN can be attributed to its ability to apply an effective regularization on the classification loss. Generating a classification map starting from only a few labeled pixels is a highly ill-posed problem, i.e., as long as the classifier correctly identifies the species on the pixels for which the ground-truth information is available, one would obtain a high accuracy regardless of what the classifier does on the unlabeled pixels, potentially producing an unrealistic classification overall (see the SGL output in Fig.~\ref{fig:fg_images}(c)). However, our prior knowledge about the species distribution dictates that the unlabeled pixels cannot be labeled arbitrarily and must satisfy certain smoothness conditions. This prior information is encoded by using a superpixel graph and adding a penalty that encourages the classification map to vary smoothly over this graph. The high expressive capacity of a fully-connected neural network, combined with this strong regularization, allows GRNN to achieve the right balance between accuracy over pixels with known labels and smoothness of classification over a large number of unlabeled pixels.
\begin{table}[h]
\centering
\caption{\small{SVM and SGL vs. GRNN (OA, $\kappa$) on the FG data with  10\%, 30\%, 50\% and 70\% of the available labeled pixels.  Classification maps for all results in Table III are presented in Figure 2 of the supplementary document. Despite higher pixelwise accuracy, SGL suffers from the propensity of generating an oversmoothed and unrealistic classification, while SVM produces a classification map that does not faithfully capture the tree crowns. (c.f. Fig. \ref{fig:fg_images}).}}
\begin{tabular}{ccccc}
\hline
& 10\% & 30\% & 50\% & 70\% \\
\hline
SVM & 73.99\%, 0.65 & 76.22\%, 0.68& 76.78\%, 0.69& 77.44\%, 0.70\\
SGL & 92.50 \%, 0.90 & 95.61\%, 0.94 & 96.05\%, 0.94 & 96.26\%, 0.95\\
GRNN &84.62\%, 0.79& 89.73\%, 0.86& 91.45\%, 0.89 & 90.56\%, 0.88\\
\hline
\end{tabular}
\label{tab:SVM-GRNN-FGdata}
\end{table}
\begin{figure*}[ht]
\centering
\subfloat[\scriptsize{ground-truth}]{
\includegraphics[width=0.195\linewidth]{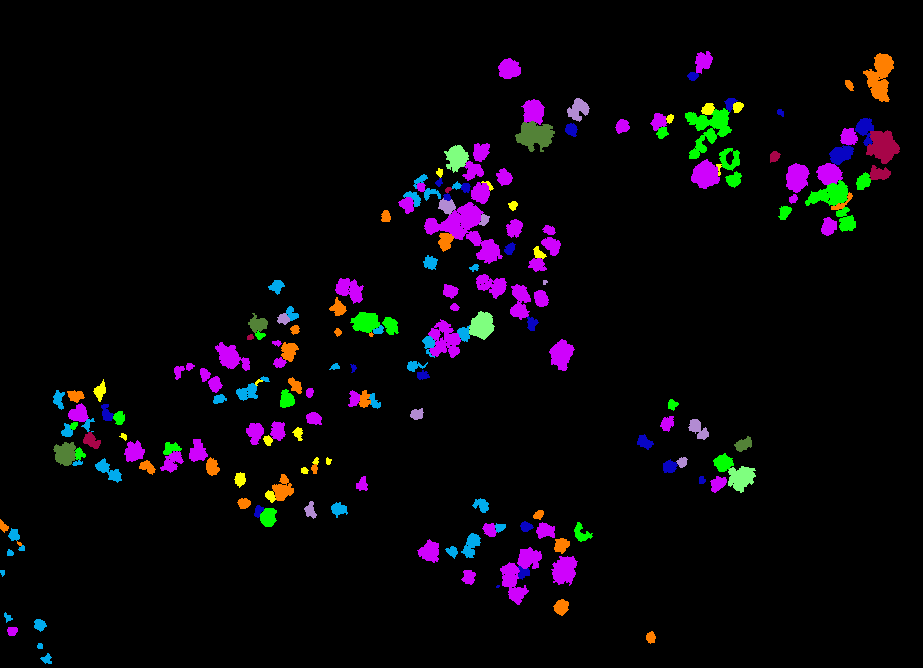}}
\subfloat[\scriptsize{SVM: $\kappa$ = 0.69, OA = 76.78\%}]{
\includegraphics[width=0.195\linewidth]{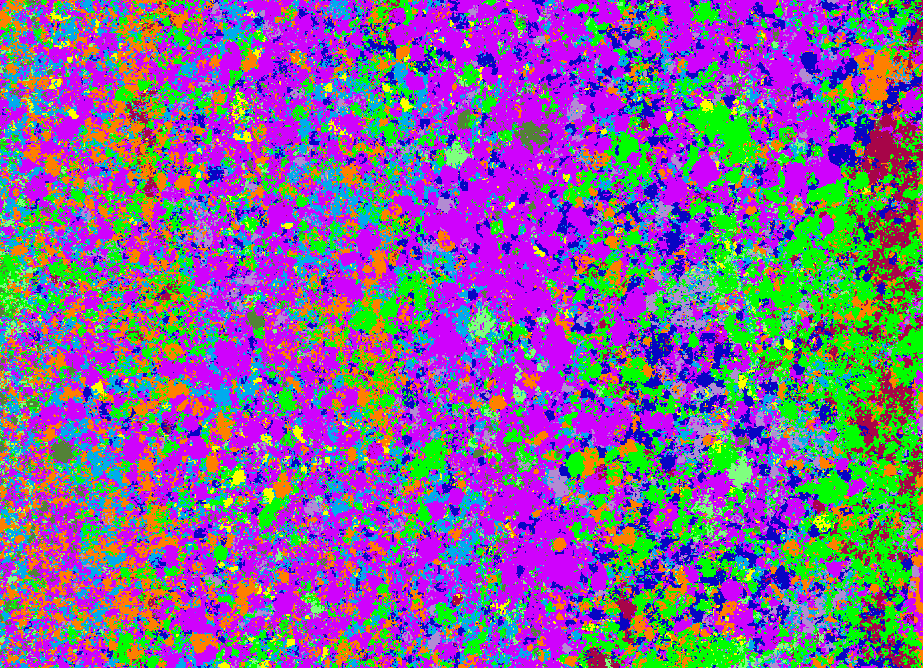}}
\subfloat[\scriptsize{SGL: $\kappa$ = 0.94, OA = 96.05\%}]{
\includegraphics[width=0.195\linewidth]{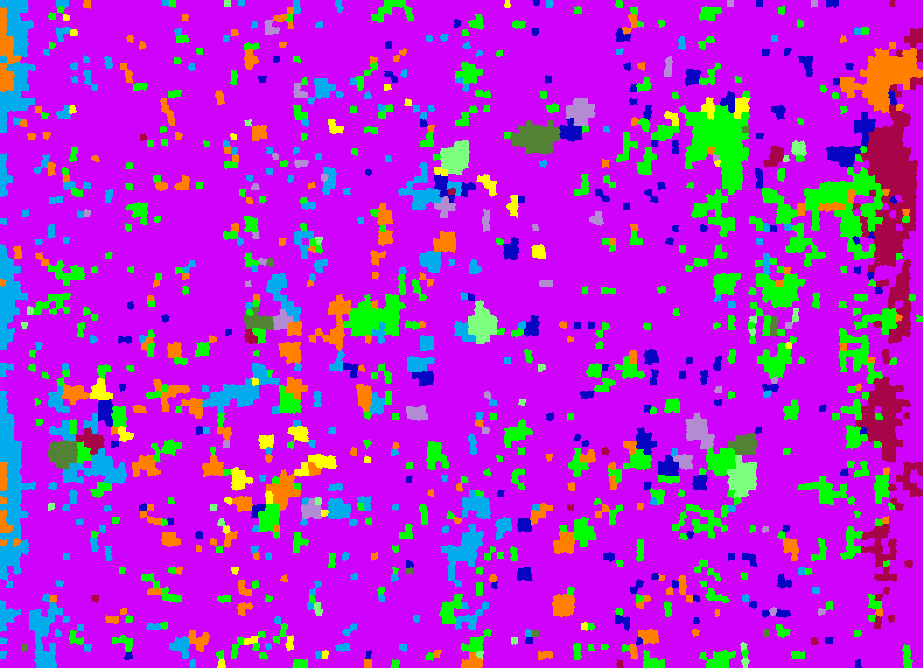}}
\subfloat[\scriptsize{GRNN: $\kappa$ = 0.89, OA = 91.45\%}]{
\includegraphics[width=0.195\linewidth]{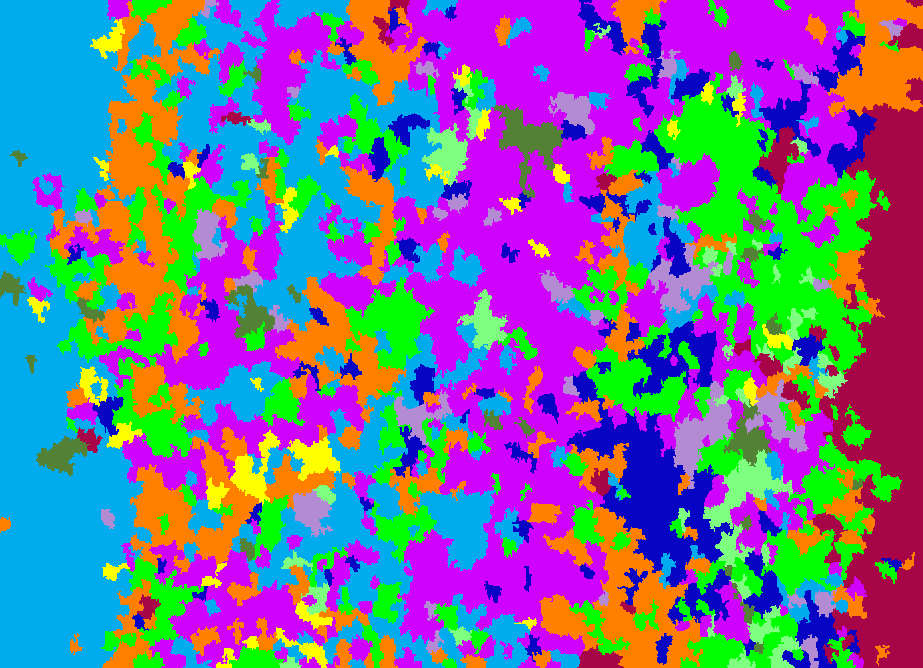}}
\subfloat{
\includegraphics[width=0.12\linewidth]{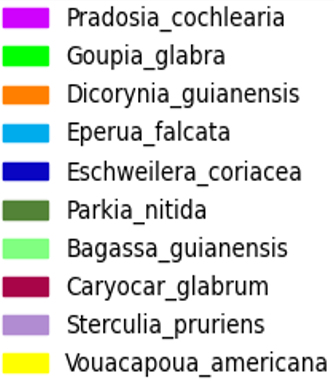}}

\caption{\small{SVM, SGL and GRNN applied to the FG data: (a) shows the labeled tree species. All three classification maps are shown: (b) SVM (c) SGL and (d) GRNN, with 50\% of the available labeled pixels used for training. Although SGL attains a higher classification accuracy, it overwhelmingly assigns the most frequent species in the training data to almost all unlabeled pixels.}}
\label{fig:fg_images}
\end{figure*}


\section{Conclusions}
We propose a novel algorithm based on superpixel graph energy minimization and a NN classifier for HSI pixel classification. The algorithm is validated on the standard Indian Pines region and shows an improved performance (varying between 6\% and 16\% increase in OA) compared to other state-of-the-art semi-supervised methods. The proposed GRNN method is also applied to a new high-resolution highly heterogeneous HSI data set and is consistent in performance (around 85\% OA) using just 10\% of the training samples (which is $<1\%$ of the total pixels). GRNN can learn from a limited labeled data set by optimizing the neural network accuracy with a superpixel graph-based energy that imposes a strong prior on the predicted classification map. In addition to the enhanced accuracy, the variance in the estimates is also significantly less compared to other methods. The reduced variance highlights the robustness of the GRNN algorithm with respect to the randomly selected samples for training. On the new, sparsely annotated, and highly heterogeneous FG data, GRNN combines information at both pixel- and superpixel-levels, thereby achieving high pixelwise classification accuracy and accurate tree-crown identification simultaneously. In future work, we intend to explore automatic feature learning from the extracted superpixels instead of using handcrafted features to enhance the performance of GRNN. A more extensive comparison of GRNN with deep-learning-based SoTA classification methods, will also be undertaken in a future study.

\noindent \textbf{Acknowledgment}: 
The Indian Pines data is provided courtesy of D. Landgrebe, Purdue University, and NASA JPL. We thank the CIRAD-Joint Research Unit Ecology of Guiana Forests for providing airborne HSI data from French Guiana, including polygons for upper-canopy crowns with tree species annotated. The authors are grateful to the Franklinia Foundation, the Isaac Newton Trust, and EPSRC grant EP/T003553/1 for financial report.   

\section*{Supplementary material}
This supplementary document contains additional experimental results that could not be accommodated in the main body of the paper. 

\begin{figure}[h]
\centering
\subfloat{
\includegraphics[width=0.70\linewidth]{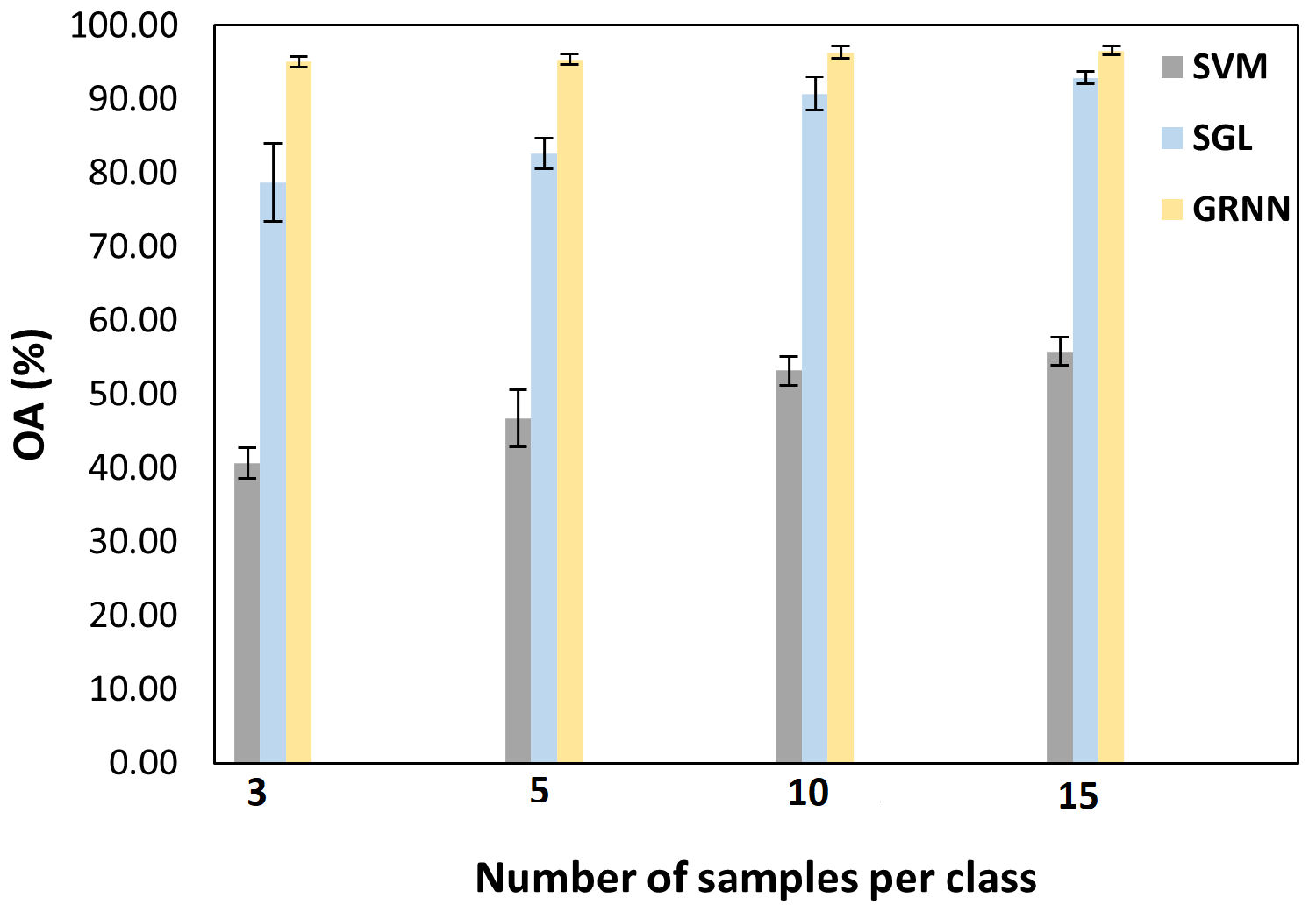}}
\caption{Comparison of the OA of SVM, SGL and the proposed GRNN-based classification on IP data. The error bar represents SD in accuracy after ten iterations on the test samples. GRNN is more robust with higher accuracy as compared to SGL.}
\label{fig:pines_histogram}
\end{figure}

In the first experiment, we compare the overall accuracy (OA) of GRNN with that of SVM (pixel-wise classifier) and SGL (superpixel-level classifier) on the Indian Pines (IP) data for different numbers of labeled pixels per class in training (see Fig. \ref{fig:pines_histogram}). GRNN attains a higher classification accuracy over both SVM and SGL and shows little variance in the overall classification accuracy over 10 random trials. \newline \newline

In Figure \ref{fig:fg_images}, we show a comparison of GRNN with SVM and SGL, two representative pixel- and superpixel-based classification algorithms, for different fractions of the available labeled pixels used for learning. Notably, the choice for SGL as a competing method is motivated by the fact that SGL was found to be the closest competitor of GRNN on the Indian Pines data.
\begin{figure*}[ht]
\centering
\subfloat[\small{RGB image}]{
\includegraphics[width=0.27\linewidth]{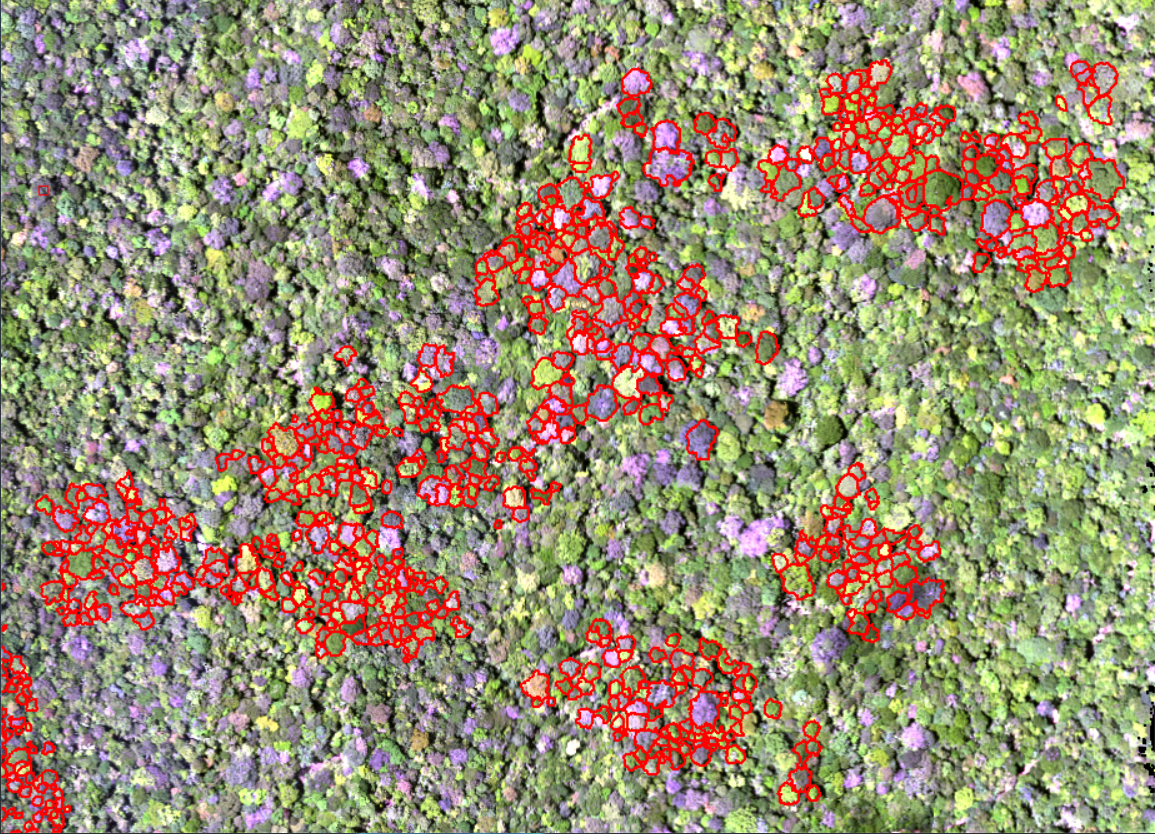}}
\subfloat[\small{ground-truth}]{
\includegraphics[width=0.27\linewidth]{FG_images/ground_truth.png}}
\subfloat{
\includegraphics[width=0.175\linewidth]{FG_images/legend.png}}
\medskip

\subfloat[\small{$\text{SVM}_{10\%}$}]{
\includegraphics[width=0.25\linewidth]{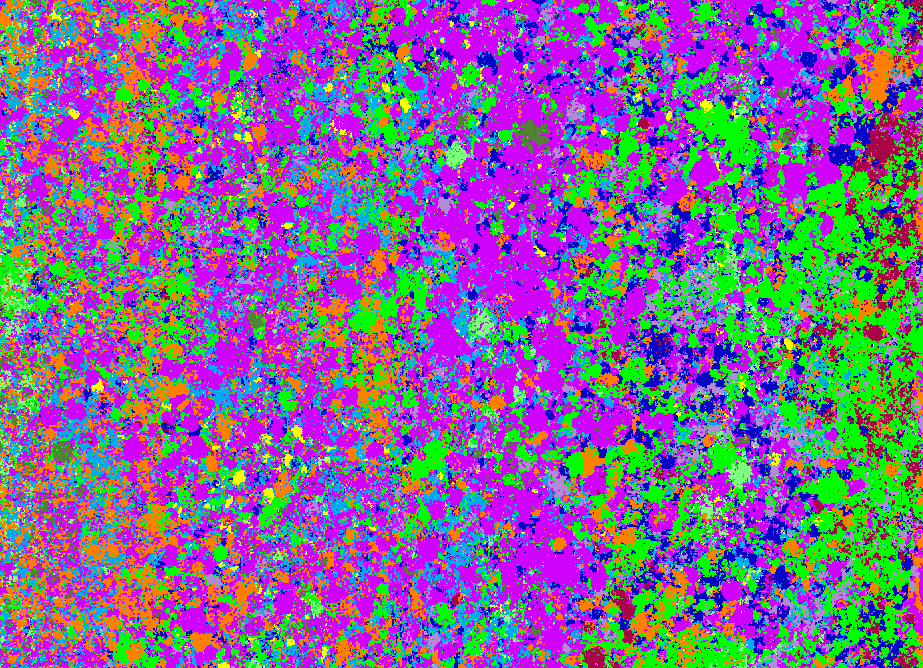}}
\subfloat[\small{$\text{SVM}_{30\%}$}]{
\includegraphics[width=0.25\linewidth]{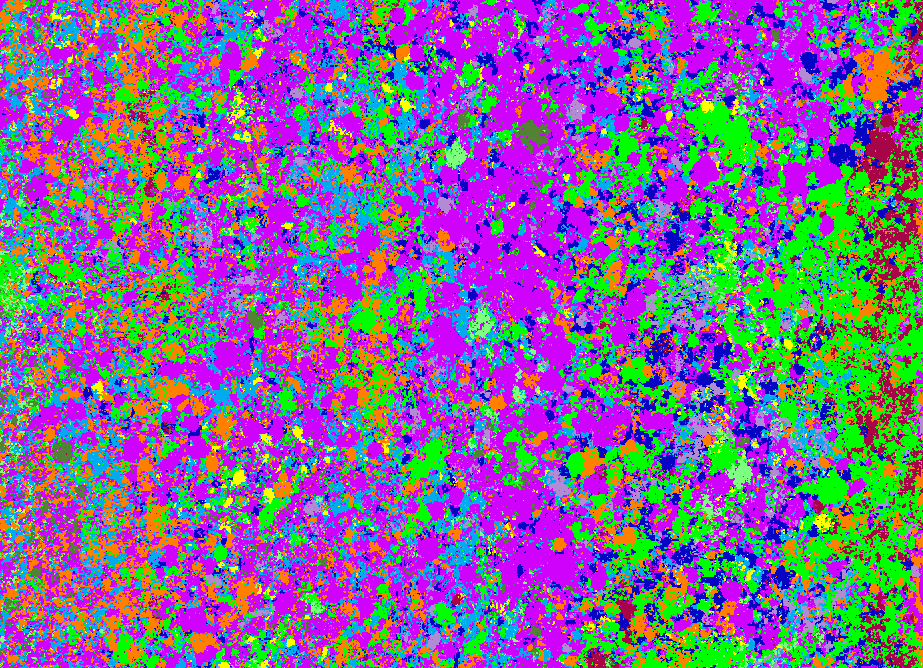}}
\subfloat[\small{$\text{SVM}_{50\%}$}]{
\includegraphics[width=0.25\linewidth]{FG_images/predicted_classes_svm_nsamp=50_fg.png}}
\subfloat[\small{$\text{SVM}_{70\%}$}]{
\includegraphics[width=0.25\linewidth]{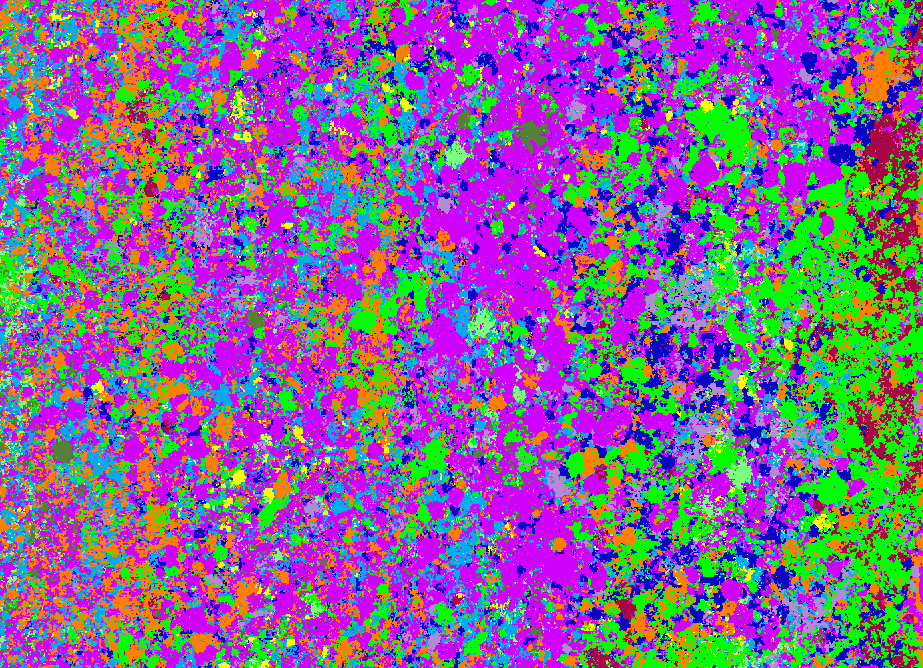}}
\medskip

\subfloat[\small{$\text{SGL}_{10\%}$}]{
\includegraphics[width=0.25\linewidth]{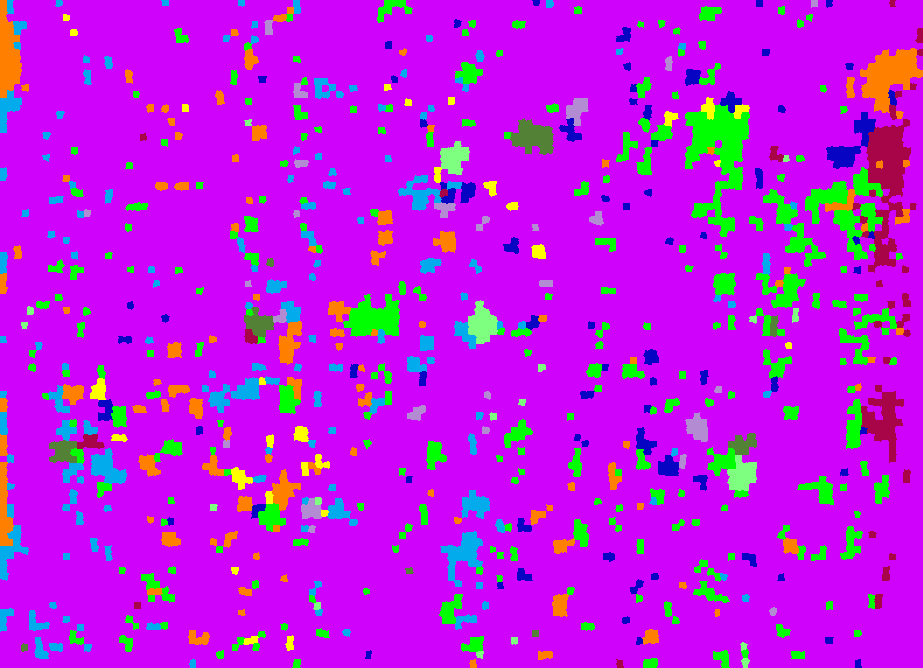}}
\subfloat[\small{$\text{SGL}_{30\%}$}]{
\includegraphics[width=0.25\linewidth]{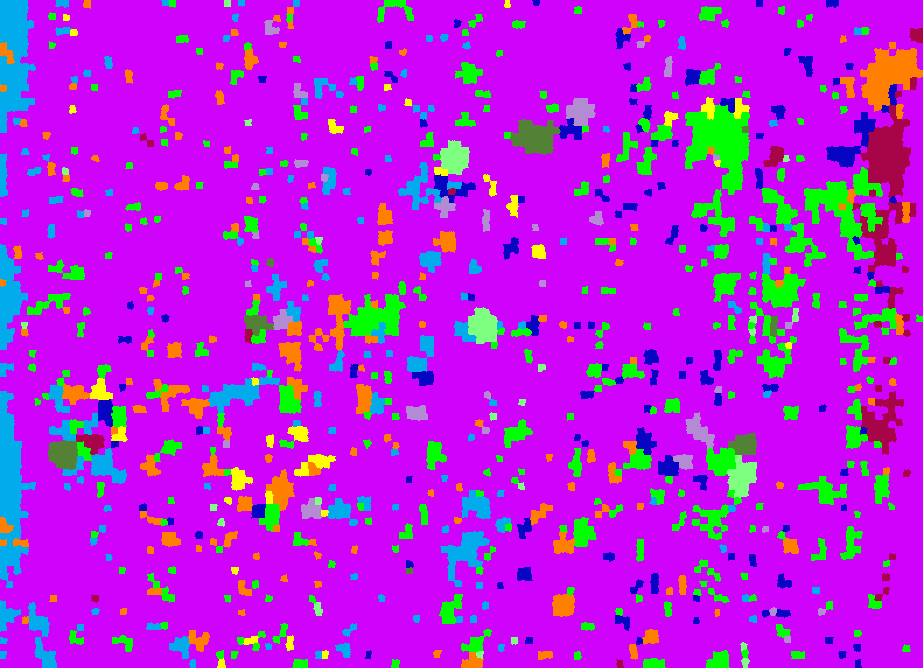}}
\subfloat[\small{$\text{SGL}_{50\%}$}]{
\includegraphics[width=0.25\linewidth]{FG_images/hms_fgdata_mosaic_50pc.png}}
\subfloat[\small{$\text{SGL}_{70\%}$}]{
\includegraphics[width=0.25\linewidth]{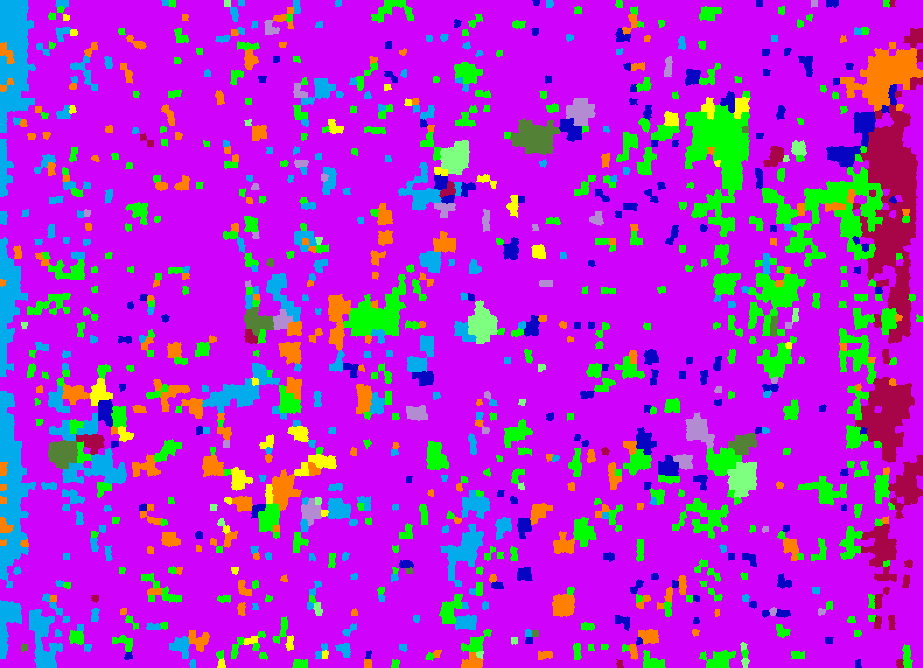}}
\medskip

\subfloat[\small{$\text{GRNN}_{10\%}$}]{
\includegraphics[width=0.25\linewidth]{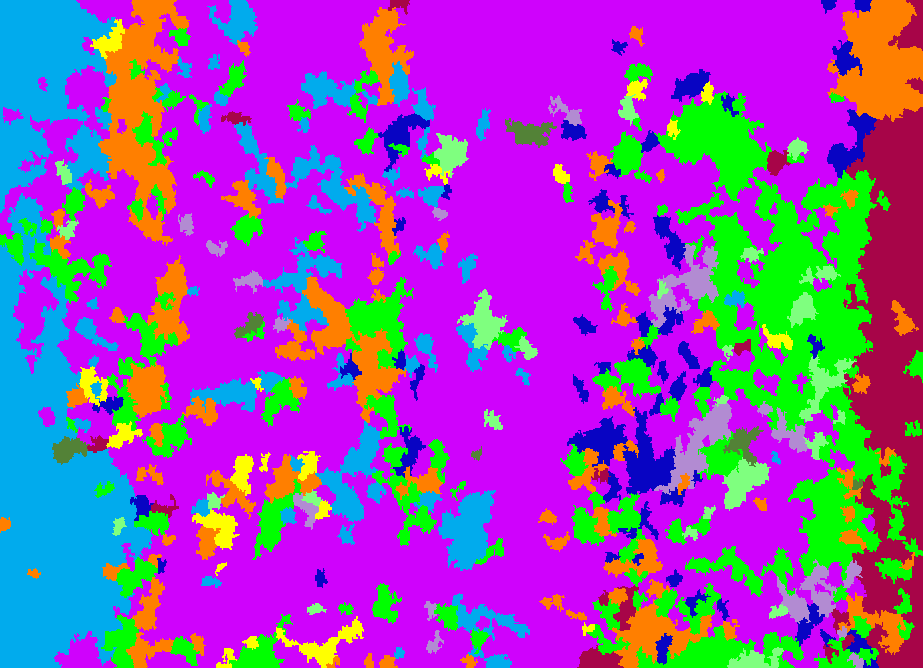}}
\subfloat[\small{$\text{GRNN}_{30\%}$}]{
\includegraphics[width=0.25\linewidth]{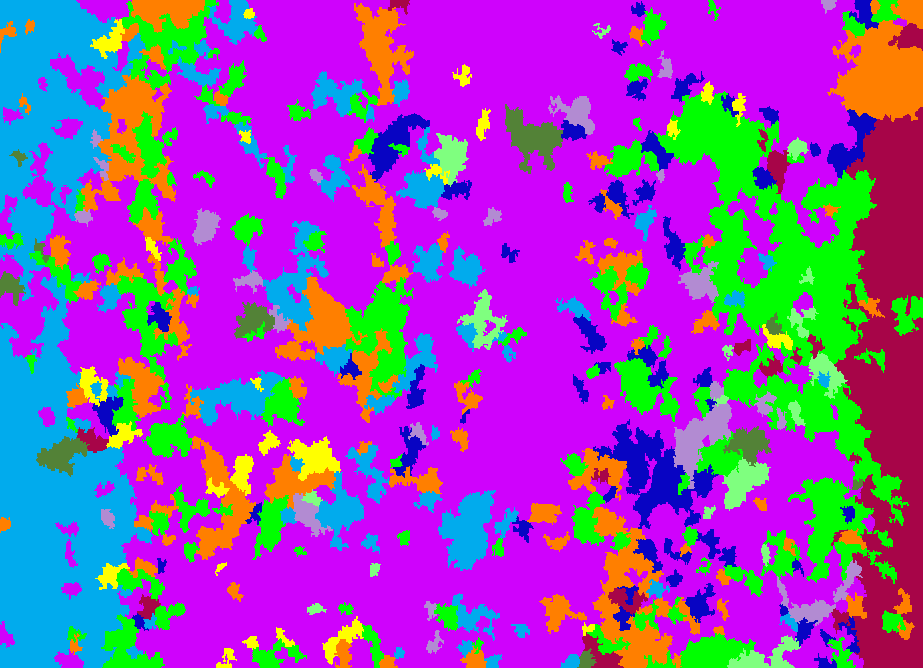}}
\subfloat[\small{$\text{GRNN}_{50\%}$}]{
\includegraphics[width=0.25\linewidth]{FG_images/predicted_classes_grnn_smoothed_nsamp=50_fg.png}}
\subfloat[\small{$\text{GRNN}_{70\%}$}]{
\includegraphics[width=0.25\linewidth]{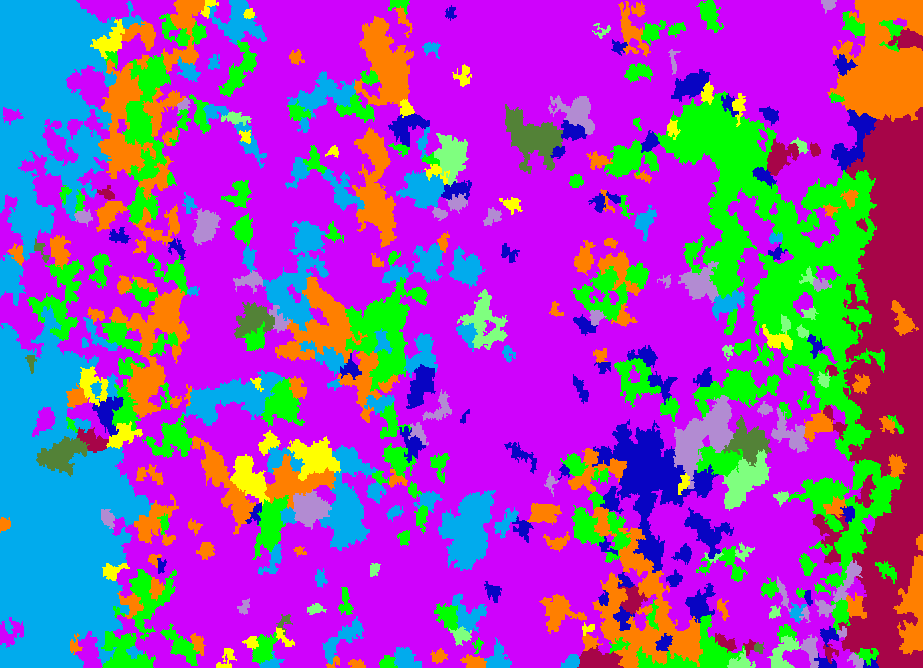}}
\medskip

\caption{\small{(a) shows the composite RGB image of the forest region and the red polygons are the annotated tree crowns for the ten most frequent species, (b) shows the labeled tree species for which the key is provided in the right corner of the image. The subscripts 10, 30, 50 and 70 for all the three methods: SVM, SGL and GRNN, correspond to 10\%, 30\%, 50\% and 70\% of the labeled  pixels used for training. The corresponding overall accuracy and $\kappa$ values for each of the clssification maps are mentioned in Table III.}}
\label{fig:fg_images}
\end{figure*}
\ifCLASSOPTIONcaptionsoff
  \newpage
\fi

%
\small{
\bibliographystyle{IEEEtranN}
\bibliography{references}
}

\end{document}